\relax
\documentclass[letterpaper]{article} 
\usepackage{aaai18}  
\usepackage{times}  
\usepackage{helvet}  
\usepackage{courier}  
\usepackage{url}  
\usepackage{graphicx}  
\usepackage{graphicx}
\usepackage{subfigure}
\usepackage{amsmath,bm}
\usepackage{array}
\usepackage{booktabs}       
\usepackage{amsfonts}       
\usepackage{nicefrac}       
\usepackage{microtype}      
\usepackage{color}      
\usepackage{booktabs}
\usepackage{multicol}
\usepackage{multirow}
\usepackage{threeparttable}
\newcommand{\tabincell}[2]{\begin{tabular}{@{}#1@{}}#2\end{tabular}}

\begin{document}
%
\title{A New Hybrid-parameter Recurrent Neural Networks for Online Handwritten Chinese Character Recognition}
\author{Haiqing Ren    Weiqiang Wang\\
University of Chinese Academy of Sciences, CAS, Beijing, China\\
\textit{Email:renhaiqing14@mails.ucas.ac.cn, wqwang@ucas.ac.cn}\\
}
\maketitle
\begin{abstract}
The recurrent neural network (RNN) is appropriate for dealing with temporal sequences. In this paper, we present a deep RNN with new features and apply it for online handwritten Chinese character recognition. Compared with the existing RNN models, three innovations are involved in the proposed system. First, a new hidden layer function for RNN is proposed for learning temporal information better. we call it Memory Pool Unit (MPU). The proposed MPU has a simple architecture. Second, a new RNN architecture with hybrid parameter is presented, in order to increasing the expression capacity of RNN. The proposed hybrid-parameter RNN has parameter changes when calculating the iteration at temporal dimension. Third, we make a adaptation that all the outputs of each layer are stacked as the output of network. Stacked hidden layer states combine all the hidden layer states for increasing the expression capacity. Experiments are carried out on the IAHCC-UCAS2016 dataset and the CASIA-OLHWDB1.1 dataset. The experimental results show that the hybrid-parameter RNN obtain a better recognition performance with higher efficiency (fewer parameters and faster speed). And the proposed Memory Pool Unit is proved to be a simple hidden layer function and obtains a competitive recognition results.
\end{abstract}

\section{Introduction}
Decades ago, RNN was proposed as a perception tool for sequence processing. With the widespread use of RNNs, many improvements have been obtained, particularly in avoiding vanishing gradient problem~\cite{cit20}. LSTM~\cite{cit14} and GRU~\cite{cit19} are two famous and popular improvements of RNNs' hidden layer function and have been applied in many tasks, such as speech recognition~\cite{cit13}, Natural Language Processing (NLP)~\cite{cit23}, character recognition~\cite{cit22}, machine translation~\cite{cit19}. Alex Graves~\cite{cit12} proposed a sequence generator based on the RNNs equipped with LSTM. He and Tang ~\cite{cit34} applied LSTM-RNN to scene text recognition. Zhang et al.~\cite{cit28} proposed a Chinese character recognizer based on RNNs equipped with LSTM and GRU. GRU was first proposed for machine translation by Bengio~\cite{cit19}. Compared with LSTM, GRU has a simpler structure and obtains competitive training results.
\par
Despite the tremendous advances and successful applications, there still remain big challenges, particularly, the hidden layer function and network structure. The hidden layer function of RNNs is an important subject and needs to be explored in depth. Compared with GRU, LSTM has a more clear and reasonable workflow but more complex structure. Compared with LSTM, GRU has a simpler structure but slightly unclear workflow. Therefore, researchers need to explore whether there are other methods for avoiding vanishing gradient problem, rather than always apply the existing methods.
\begin{figure}[ptbh]
\centering
\includegraphics[width=3in]{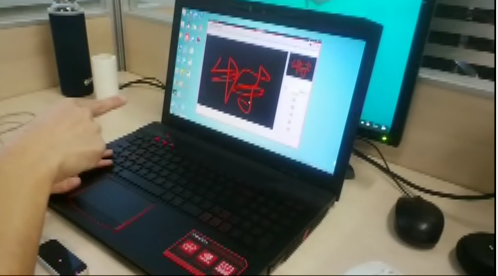}\newline\caption{The Example of In-air Handwritten with the Leap Motion Sensor.}%
\label{fig:1}%
\end{figure}
In-air hidden written is a new kind of human-computer interaction way. With some sensors (e.g., the Leap Motion sensor) and computers, people write in-air and computer recognizes what you write quickly. This kind of human-computer interaction way is shown in Fig.~\ref{fig:1}. Generally speaking, in-air handwritten Chinese characters recognition is more challenge than traditional handwritten Chinese character recognition (HCCR)~\cite{cit1,cit3,cit4}. First, each character has only one stroke without any mask of pen up-and-down. Second, in-air handwritten strokes would be more squiggly lines than handwritten strokes on touch screen. Fig.~\ref{fig2} gives some examples to distinguish the difference between IAHCC and HCC on touch screen.
\begin{figure}[htbp]
  \centering
  \subfigure[In-air handwritten Chinese characters]{
  \centering
    \includegraphics[width=0.19\textwidth]{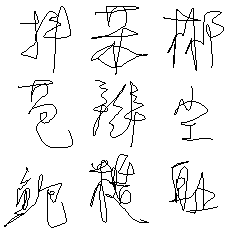}
  \label{fig2:side:a}
  }
  \subfigure[Handwritten Chinese characters on touch screen]{
  \centering
  \includegraphics[width=0.19\textwidth]{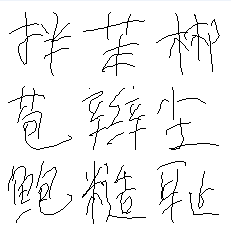}
  \label{fig2:side:b}
  }
  \caption{Comparison of HCC and IAHCC~\cite{cit31}.}
  \label{fig2}
  \end{figure}
Since in-air handwritten is such an amazing way of human-computer interaction, many researchers explore in this filed. Qu \emph{et al.}~\cite{cit27} presented a multi-stage classifier for IAHCCR, and their system achieved a relatively low accuracy. Qu \emph{et al.}~\cite{cit24} presented a new feature representation to extend the power of 8-direction feature and applied it in IAHCC recognition. 8-direction-feature had been shown to be a discriminative feature in many works and achieved a good performance~\cite{cit3,cit5,cit6}. Ren~\emph{et al.}~\cite{cit31} proposed an end-to-end recognizer for OIAHCCR based on a new RNN. In Ren's work, it is the first time for RNN to be used for OIAHCCR and to obtain a high recognition accuracy.

In the proposed system, three contributions are proposed to increase the recognition accuracy or the calculation speed of the in-air handwritten Chinese character RNN recognizer.
\begin{itemize}
\item First, a new hidden layer function MPU is proposed. Compared with LSTM and GRU respectively, the proposed MPU has fewer parameters and a straightforward workflow. A series of experiments were carried out on IAHCC-UCAS2016 dataset. It is proved that the proposed hidden layer function (MPU) obtains a high recognition accuracy.
\end{itemize}

\begin{itemize}
\item Second, a hybrid-parameter RNNs architecture is proposed. Compared with general RNNs and bidirectional RNNs correspondingly, the proposed hybrid-parameter RNNs architecture obtains competitive recognition accuracy with fewer parameters and faster calculation speed.
\end{itemize}

\begin{itemize}
\item Third, when all the size of hidden layers of a RNN are same, we make a suggestion that all the outputs of each layer are stacked as the output of network. By synthesizing all the outputs of each layer, stacked method increases the recognition accuracy without increasing the parameters.
\end{itemize}
\begin{figure*}[ptbh]
\centering
\includegraphics[width=6.5in]{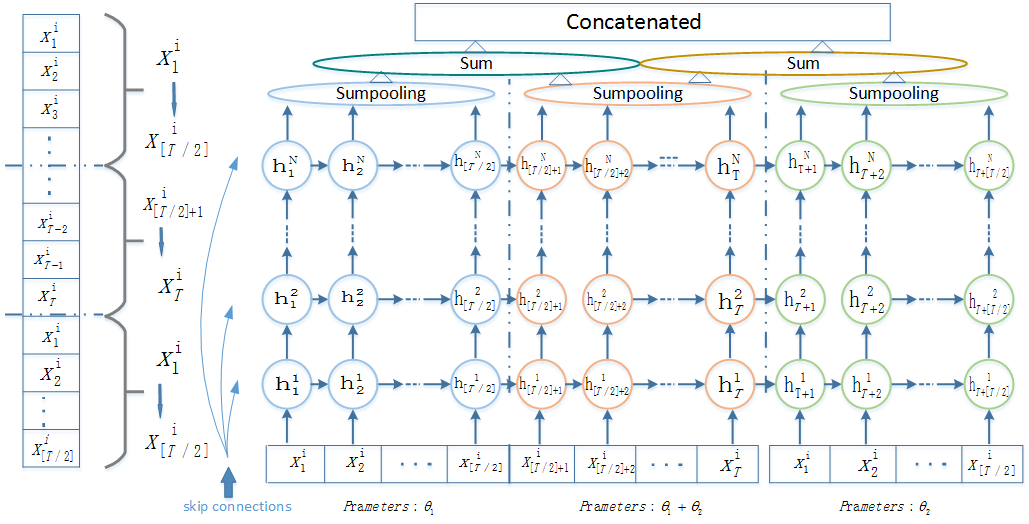}\newline\caption{Hybrid Parameter RNNs.}%
\label{fig:4}%
\end{figure*}
\section{Basic RNN System}
\subsection{Hybrid-Parameter RNNs}
Calculation speed and parameter quantity are the two significant performance index for all the neural networks. However, designing an efficiency neural network with fewer parameters and faster calculation speed is a challenging research. Fig.~\ref{fig:4} shows the proposed hybrid-parameter RNNs system and input data processing of input sample $i$. The left part of Fig.~\ref{fig:4} shows all the input character location sequence is divided into two parts. Then the new sequence of sample $i$ is generated as:
\begin{equation}\label{3}
\begin{split}
\bm{x}^i=&(\bm{x}_1^{(i)},\bm{x}_2^{(i)},\cdots,\bm{x}_{[T/2]}^{(i)},\bm{x}_{[T/2]+1}^{(i)},\bm{x}_{[T/2]+2}^{(i)},\\
&\cdots,\bm{x}_T^{(i)},\bm{x}_1^{(i)},\bm{x}_2^{(i)},\cdots,\bm{x}_{[T/2]}^{(i)}).
\end{split}
\end{equation}
As shown in Eq.~(\ref{3}), the new sequence of sample $i$ is represented as a sequence with $[T/2]+T$ location dots. The network calculation process is shown in the right part of Fig.~\ref{fig:4}. To begin with, two sets of RNN parameters ($\theta_1$ and $\theta_2$, parameters in $\theta_1$ are initialized randomly and parameters in $\theta_2$ is initialized as zeros) are initialized with the same scale. Given a character sample $i$, the new system input is a sequence of location dots, as shown in Eq.~(\ref{3}). When the RNN iterating on temperal dimension, the parameters are changed as follows. During the first $[T/2]$ time step's iteration, the RNN calculates with parameters $\theta_1$. Then the parameters change to $\theta_1 + \theta_2$ when the RNN iterates the time steps from $[T/2]$ to $T$. The RNN calculates the iteration of the last $[T/2]$ time step with the only parameters $\theta_2$. For each time step of first $[T/2]$ time, the hidden layer states are computed by%
\begin{equation}\notag
\bm{h}_t^1=H_1(\bm{x}_t,\bm{h}_{t-1}^1; \bm{\theta}_1^{1})%
\end{equation}
\begin{equation}\notag
\bm{h}_t^n=H_n(\bm{x}_t, \bm{h}_t^{n-1}, \bm{h}_{t-1}^n;\bm{\theta}_1^{n}),
\end{equation}
\begin{equation}\label{5}
where\ t=1,2,\cdots,[T/2].\ n=1,2,\cdots,N
\end{equation}
where $H_1$ and $H_n$ denote the first and $n$th hidden layer hidden layer function respectively. $\bm{\theta}_1^{1}$ and $\bm{\theta}_1^{n}$ denote the corresponding network parameters. $N$ denotes that there are $N$ hidden layers. For each time step of second $T-[T/2]$ time, the hidden layers states are computed by%
\begin{equation}\notag
\bm{h}_t^1=H_1(\bm{x}_t,\bm{h}_{t-1}^1; \bm{\theta}_1^{1}+\bm{\theta}_2^{1})%
\end{equation}
\begin{equation}\notag
\bm{h}_t^n=H_n(\bm{x}_t, \bm{h}_t^{n-1}, \bm{h}_{t-1}^n;\bm{\theta}_1^{n}+\bm{\theta}_2^{n}),
\end{equation}
\begin{equation}\label{7}
where\ t=[T/2]+1,[T/2]+2,\cdots,T. \ n=1,2,\cdots,N,
\end{equation}
where $\bm{\theta}_1^{n}+\bm{\theta}_2^{n}$ denotes the network parameters of the $n$th layer during the second $T-[T/2]$ time step. For each time step of the last $[T/2]$ time, the hidden layers states are computed by%
\begin{equation}\notag
\bm{h}_t^1=H_1(\bm{x}_t,\bm{h}_{t-1}^1; \bm{\theta}_2^{1})%
\end{equation}
\begin{equation}\notag
\bm{h}_t^n=H_n(\bm{x}_t, \bm{h}_t^{n-1}, \bm{h}_{t-1}^n;\bm{\theta}_2^{n}),
\end{equation}
\begin{equation}\label{9}
where\ t=T+1,T+2,\cdots,T+[T/2]. \ n=1,2,\cdots,N,
\end{equation}
where $\bm{\theta}_2^{n}$ denotes the network parameters of $n$th layer during the last $[T/2]$ time steps. During the whole temporal iteration, both $\bm{\theta}_1^{1}$ and $\bm{\theta}_1^{1}$ are participated in the RNN calculation with all the location dots of the original sample $i$. After all the $T+[T/2]$ time step iterations, $T+[T/2]$ hidden layer states are generated at the $N$th layer, e.g., $\bm{h}_1^N,\bm{h}_2^N,...\bm{h}_t^N...\bm{h}_{T+[T/2]}^N$. Then the final output of the RNN is computed by
\begin{equation}\label{12}
{\bm{y}=\bm{b}_y+ (\bm{W}_{h_{U_1} y} U_1+\bm{W}_{h_{U_2} y} U_2 })
\end{equation}
\begin{equation}\label{10}
U_1=\sum_{t=1}^{T} \bm{h}_t^N
\end{equation}
\begin{equation}\label{11}
U_2=\sum_{t=[T/2]+1}^{T+[T/2]} \bm{h}_t^N
\end{equation}
where ${\bm{y}}$ denotes the output vector, ${\bm{W}_{h_{U_1} y}}$ and ${\bm{W}_{h_{U_2} y}}$ denote the weight matrix from $U_1$ and $U_2$ to the fully-connected layer (output layer). ${\bm{b}_y}$ denotes the fully-connected layer bias vector. The sum operation (sum-pooling in Eq.~(\ref{10}) and Eq.~(\ref{11})) proposed by Ren~\emph{et al.}~\cite{cit31}. Then a softmax regression is used on the fully-connected layer outputs for computing the class probability distribution.
Since an in-air Chinese handwritten character generally corresponds to a long sequence of dot locations and the proposed network architecture involves five hidden layers, we add the \emph{skip connections}~\cite{cit12} from the input layer to all hidden layers, and from all hidden layers to the fully-connected layer, to alleviate the \emph{vanishing gradient} problems~\cite{cit20}.
\subsubsection{Computing Speed and Parameters Quantity}
General RNNs are computed with only one set of parameters during all the iteration at temperal dimension. The parameters are changed during iterating at temporal dimension in the proposed RNN with hybrid parameters. Experimental results show the proposed RNN structure obtains a higher recognition accuracy with a smaller hidden layer size. The smaller hidden layer size means the fewer parameters, e.g., a general RNN with five 256-size hidden layers could obtain a recognition accuracy of 92.6\%. The recognition accuracy can not be increased when the hidden layer size increases. However, recognition accuracy would be decreased when the hidden layer size decreases. The proposed RNN with hybrid parameters could obtain a higher recognition accuracy of 92.9\% with 128-size. The numerical relationship between the two kinds of RNN could be represented as:
\begin{equation}\label{13}
\begin{split}
Q=&3*(\underbrace {D^1*D^2+D^2*D^3+\cdots+D^{N-1}*D^{N}}_{W_1})\\
&+3*(\underbrace {D^1*D^1+D^2*D^2+\cdots+D^{N}*D^{N}}_{W_2})
\end{split}
\end{equation}
\begin{equation}\label{14}
\begin{split}
q=&2*(3*(\underbrace {d^1*d^2+d^2*d^3+\cdots+d^{N-1}*d^{N}}_{w_1})\\
&+3*(\underbrace {d^1*d^1+d^2*d^2+\cdots+d^{N}*d^{N}}_{w_2}))
\end{split}
\end{equation}
where $Q$ denotes the parameter quantity of the general RNN. $D^n$ denotes the hidden layer size of $n$th hidden layer. $W_1$ denotes the parameters of all the weighted matrix of each layer inputs. $W_2$ denotes the parameters of all the weighted matrix from the hidden layer state $h_{t-1}$ to $h_{t}$. And $q$, $d^n$, $w_1$, $w_2$ have the similar definition for the proposed hybrid-parameter RNN. The multiplicator $3$ in Eq.~(\ref{13}) and Eq.~(\ref{14}) represents the number of state vectors in the hidden layer function. The hidden layer function used in RNN is GRU. Therefore, the number of state vectors is $3$. The multiplicator $2$ in Eq.~(\ref{14}) represents there are two set of parameters with the same scale. From the Eq.~(\ref{13}) and Eq.~(\ref{14}) we can see that $Q$ is as twice as $q$  the hybrid-parameter RNN and the general RNN have same hidden layer size.
\par
Compared with the bidirectional RNN, the hybrid-parameter RNN is calculated with only $T+[T/2]$ time steps' iteration, while the bidirectional RNN needs to be calculated with $2*T$ time steps' iteration. Therefore, the proposed hybrid-parameter RNN has a faster computing speed than the bidirectional RNN and obtains a competitive result.
\subsection{Learning Parameters}
At the top of the RNN, a softmax layer is used to generate the probability distribution corresponding to 3873 character classes. To train the neural network, the following loss function is widely used and minimized, i.e.,
\begin{equation}\label{15}
\begin{split}
&J(\bm{\theta}_1,\bm{\theta}_2)=\\
&-\frac{1}{m} [\sum_{i=1}^m \sum_{l=1}^K 1\{C^{(i)}=l\}log \frac{exp(y_l^{(i)}(\bm{\theta}_1,\bm{\theta}_2))}{\sum_{j=1}^K exp(y_{j}^{(i)}(\bm{\theta}_1,\bm{\theta}_2))}]
\end{split}
\end{equation}
\begin{equation}\label{16}
1\{C^{(i)}=l\}=\left\{
 \begin{aligned}
 1 \qquad &C^{(i)}=l \\
 0 \qquad &C^{(i)}\neq l .\\
\end{aligned}
\right.
\end{equation}
where $m$ is the total number of training patterns, and the $1\{\cdot\}$ function selects the true class of the example. $y_{j}^{(i)}$ denotes the corresponding output's $j$th element of the RNN given a pattern $\bm{x}^{(i)}$.
The minimization of the loss function corresponds to maximizing the probability of correct classification of patterns in essence. Then the optimal parameters of the proposed system are obtained by constantly updating parameters based on the rmsprop~\cite{cit21} method, a form of stochastic gradient descent.
\subsection{Memory Pool Unit}
\begin{figure}[ptbh]
\centering
\includegraphics[width=3.2in]{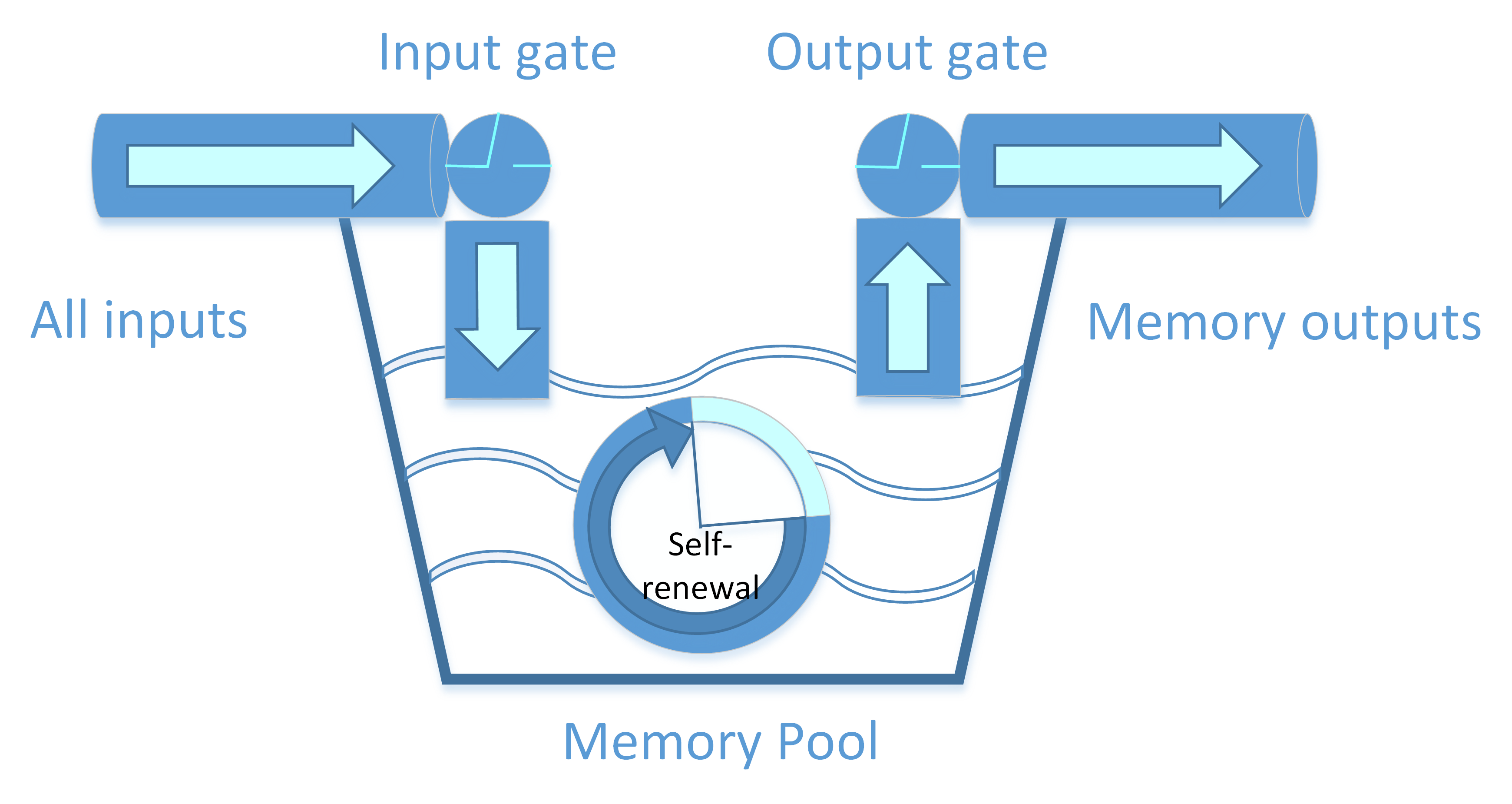}\newline\caption{Memory Pool Unit.}%
\label{fig:5}%
\end{figure}
Hidden layer function is significant for RNNs. LSTM and GRU are the two popular hidden layer functions used for many tasks. As a hidden layer function with a simpler architecture and fewer parameters, GRU obtains similar performance with LSTM. To make computation and implementation much simpler, memory state is wiped out in the GRU. However, hidden layer function with memory state make the whole hidden layer state calculation process more reasonable and convictive. we propose a new type of hidden layer function with memory state that has been motivated by the LSTM and GRU but is simpler than LSTM. In this section, we focus on describing the proposed hidden layer function Memory Pool Unit (MPU).
MPU is based on a simple and more straightforward hidden layer state calculation process. As shown in Fig.~\ref{fig:5}, the core of the proposed hidden layer function is a memory pool. There are two gates in the architecture of MPU. One is input gate and the other is output gate. The memory pool stores a state vector, which is updated by the inputs with the limitation of input gate. With the control of output gate, the new hidden layer state is generated by using the updated memory pool state. The proposed MPU could be described as follows,
\begin{equation}\label{17}
{\bm{i}_t=\sigma(\bm{W}_{xi} \bm{x}_t+\bm{W}_{hi} \bm{h}_{t-1}+\bm{b}_i)},
\end{equation}
\begin{equation}\label{18}
 \begin{split}
 \bm{m}_t=tanh(\bm{W}_{xm} (\bm{i}_t \odot \bm{x}_t)+\bm{W}_{hm}( \bm{i}_t \odot \bm{h}_{t-1}))\\
          +(\bm{1}-\bm{i}_t)\odot \bm{m}_{t-1}
 \end{split}
\end{equation}
\begin{equation}\label{19}
{\bm{o}_t=\sigma(\bm{W}_{xo} \bm{x}_t+\bm{W}_{ho} \bm{h}_{t-1}+\bm{b}_o)},
\end{equation}
\begin{equation}\label{20}
{\bm{h}_t=\bm{o}_t\odot \bm{m}_{t}}
\end{equation}
where $\bm{i}$ and $\bm{o}$ denote \emph{input gates} and \emph{output gates}. $\bm{m}$ denotes the memory pool state. Eq.~(\ref{18}) shows that all the inputs of memory pool are restricted by the same input gate. The inputs include the last layer's outputs, the network inputs and the hidden layer state of last time step ($\bm{h}_{t-1}$). The last layer's outputs and the network inputs are represented by $\bm{x}_t$ in equations from Eq.~(\ref{17}) to Eq.~(\ref{19}). As shown in Eq.~(\ref{18}), when the input gate is close to zero, the memory pool state is forced to ignore the previous inputs. Therefore, $\bm{m_t}$ keeps its former state. For other situations, the memory pool state is updated by the inputs and its former state. The current hidden layer state is calculated using the memory pool state under the control of the output gate. The whole processing mechanism is like a pool with two valves and a self-renewal capacity. The input gate allows useful information to pour into the pool, then the pool state self-renews, and the useful information drains out through the output gate at last. The self-renewal capacity of memory pool state, as shown in Eq.~(\ref{18}), is the most important part of the MPU. On one side, the input gate restricts not only the last time step hidden layer state but also the current inputs. By restricting the hidden state of last time step, significant information is taken out for later computation and irrelevant information is ignored. The restriction to the current input is also important for our memory pool. Without the restriction, a great deal of useless information will be stacked in the pool when the RNNs are deep (to be exactly, without the restriction to current input, the training process of RNNs becomes difficult if there are more than two hidden layers). On the other side, the information of pool input and former memory pool state used to renew the memory pool state is complementary in a way. This kind of complementary makes better use of all the information and avoids irrespective information being stacked in the pool. Because all the inputs are restricted by the input gate, it is significant that the network inputs are used for compensating the information loss of the vertical direction.  ${W}_{xi}$ denotes the weighted matrix from input vector to input gate. $\bm{W}_{hi}$, $\bm{W}_{ho}$, $\bm{W}_{xo}$, $\bm{W}_{xm}$ and $\bm{W}_{hm}$ have similar meanings. $\bm{b}_i$ and $\bm{b}_o$ are the biases of the corresponding gates.
\subsection{Memory Pool Unit with Input Compensation}
As network input compensation is significant for the proposed MPU, another method for compensating the information loss of the vertical direction is proposed. We change the Eq.(\ref{20}) of last subsection into
\begin{equation}\label{21}
{\bm{h}_t=tanh(\bm{o}_t\odot \bm{m}_{t}}+ReLu(\bm{W}_{xc}\bm{x}_t))
\end{equation}
where $\bm{W}_{xc}$ denotes the compensation weighted matrix of last layer's outputs. With the compensation item in Eq.(\ref{21}), network input is no longer necessary for each layer except the first layer.
\subsection{Stacked Different Hidden Layer States}
In the general RNNs, the neural network outputs are calculated by using only the last hidden layer state~\cite{cit28} or all the hidden layers states~\cite{cit12}. It is better to calculate the outputs by using all the hidden layers states. Generally, the calculation of all the hidden layers states is always weighted sum, such as:
\begin{equation}\label{22}
{\sum_{n=1}^N \bm{W}_{h_{U_1}^n y} U_1^n},
\end{equation}
where ${\bm{W}_{h_{U_1}^n y}}$ $(n=1,2...n)$ denotes the weight matrix from $n$th hidden layer to the full connection layer (output layer). The weight matrix brings a great amount of parameters, which have a bad effect on neural network perception. When all the hidden layers' sizes are same, we make a adaptation that all the hidden layer states are stacked as the output of network. The calculation process could be described as Eq.(\ref{10}) and Eq.(\ref{11}).
The hidden layer neurons are calculated by the neural network parameters, the parameters change during training process. Therefore, the attributes and function of neurons are relative rather than absolute. Due to neurons' relativity, the sum calculation without weight matrix could achieve the similar goals with weighted sum calculation, which are proved as follows. Given the RNN output $\bm{y}$ with weighted sum calculation   :
\begin{equation}\label{23}
{\bm{y}=\bm{b}_y+ \sum_{n=1}^N\bm{W}_{h_{U_1}^n y} U_1^n+\sum_{n=1}^N\bm{W}_{h_{U_2}^n y} U_2^n ,}
\end{equation}
and the network output $\bm{y}$ could be represented as:
\begin{equation}\label{24}
\begin{split}
&\bm{b}_y+\sum_{n=1}^N\bm{W}_{h_{U_1}^n y} U_1^n+\sum_{n=1}^N\bm{W}_{h_{U_2}^n y} U_2^n \\
=&\bm{b}_y+\sum_{n=1}^N (\bm{W}_{h_{U_1}^n y} \sum_{t=1}^{T} \bm{h}_t^n+ \bm{W}_{h_{U_2}^n y}\sum_{t=[T/2]+1}^{T+[T/2]}\bm{h}_t^n)\\
=&\bm{b}_y+\sum_{n=1}^N (\sum_{t=1}^{T} \bm{W}_{h_{U_1}^n y} \bm{h}_t^n+\sum_{t=[T/2]+1}^{T[T/2]} \bm{W}_{h_{U_2}^n y}\bm{h}_t^n )\\
=&\bm{b}_y+\sum_{n=1}^N\sum_{t=1}^{T} \Gamma(\bm{x}_t, \bm{h}_t^{n-1}, \bm{h}_{t-1}^n;\Theta(\bm{\theta}_1^n,\bm{\theta}_2^n, \bm{W}_{h_{U_1}^n y})) \\
&+\sum_{n=1}^N\sum_{t=[T/2]+1}^{T+[T/2]} \Gamma(\bm{x}_t, \bm{h}_t^{n-1}, \bm{h}_{t-1}^n;\Theta(\bm{\theta}_1^n,\bm{\theta}_2^n, \bm{W}_{h_{U_2}^n y}))
\end{split}
\end{equation}
As shown in Eq.(\ref{24}), the network output $\bm{y}$ is a function expression of RNN parameter ($\bm{\theta}_1^n$, $\bm{\theta}_2^n$) and full-layer weighted matrix parameter ($\bm{W}_{h_{U_1}^n y})$, $\bm{W}_{h_{U_2}^n y})$). And the network output $\bm{y}$ of RNN with stacked different hidden layer states is computed as:
\begin{equation}\label{25}
\bm{y}=\bm{b}_y+ \bm{W}_{h_{U_1} y} U_1+\bm{W}_{h_{U_2} y} U_2,
\end{equation}
and the network output $\bm{y}$ could be represented as:
\begin{equation}\label{26}
\begin{split}
&\bm{b}_y+\bm{W}_{h_{U_1} y} U_1+\bm{W}_{h_{U_2} y} U_2 \\
=&\bm{b}_y+\sum_{n=1}^N (\bm{W}_{h_{U_1} y} \sum_{t=1}^{T} \bm{h}_t^n+ \bm{W}_{h_{U_2} y}\sum_{t=[T/2]+1}^{T+[T/2]}\bm{h}_t^n)\\
=&\bm{b}_y+\sum_{n=1}^N \sum_{t=1}^{T} \bm{W}_{h_{U_1} y}\bm{h}_t^n+\sum_{n=1}^N \sum_{t=[T/2]+1}^{T+[T/2]}\bm{W}_{h_{U_1} y}\bm{h}_t^n       \\
=&\bm{b}_y+\sum_{n=1}^N \sum_{t=1}^{T}\Gamma(\bm{x}_t, \bm{h}_t^{n-1}, \bm{h}_{t-1}^n;\Theta(\bm{\theta}_1^n,\bm{\theta}_2^n,\bm{W}_{h_{U_1} y}))\\
+&\sum_{n=1}^N \sum_{t=[T/2]+1}^{T+[T/2]}\Gamma(\bm{x}_t, \bm{h}_t^{n-1}, \bm{h}_{t-1}^n;\Theta(\bm{\theta}_1^n,\bm{\theta}_2^n,\bm{W}_{h_{U_2} y}))
\end{split}
\end{equation}
As shown in Eq.(\ref{26}), the network output $\bm{y}$ is a function expression of RNN parameter ($\bm{\theta}_1^n$, $\bm{\theta}_2^n$) and full-layer weighted matrix parameter ($\bm{W}_{h_{U_1} y}$, $\bm{W}_{h_{U_2} y}$). The full-layer weighted matrix parameter are mainly used for the integration of feature vectors learned by each layer. $\bm{y}$ calculated by weighted sum means more parameters. Difficulties are brought by the parameter increase. In a sense, Eq.~(\ref{24}) and Eq.~(\ref{26}) are mathematically equivalent due to parameter variation. The sum calculation has obvious advantages during training our system. Although the sum and weighted sum calculation could ideally achieve the same training effect, fewer parameters of sum calculation bring faster training process and higher recognition accuracy. Compared with the outputs calculated by using only the last hidden layer state, outputs calculated by using sum calculation could even obtain a higher recognition accuracy without increasing the amount of parameters. The proposed sum calculation is an optimal hidden layers state process method.

\section{Experimental Results}
All the experiments are carried out on IAHCC-UCAS2016 dataset and CASIA-OLHWDB1.1 dataset. IAHCC-UCAS2016 is a dataset of in-air handwritten Chinese characters containing 3873 character classes in total. Concretely, it contains 3811 Chinese characters, 52 case-sensitive English letters, and 10 digits. Each character class has 116 samples. For each class, we choose 92 samples as training set, and the remaining 24 samples are used as testing set. 14 samples in the training set are randomly sampled to form the validation set.
For further evaluation of the proposed methods, experiments are also carried out on CASIA-OLHWDB1.1 dataset. CASIA-OLHWDB1.1 dataset is a handwritten Chinese character dataset which contains 3755 characters from GB2312 written by 300 writers. The data from 240 persons are used for training, and the data from the remaining 60 persons are used for testing.
\begin{table*}[htbp]
\setlength{\abovecaptionskip}{10pt}
\setlength{\belowcaptionskip}{10pt}
  \centering
  \begin{threeparttable}
  \caption{Recognition accuracy of different methods}
  \label{table1}
    \begin{tabular}{ccccccccccc}
    \toprule
    \multirow{2}{*}{}&\multirow{2}{*}{Method}&
    \multicolumn{2}{c}{\tabincell{c}{General\\ RNNs\\256 }}&\multicolumn{3}{c}{\tabincell{c}{Hybrid-parameter\\ RNNs\\128}}&\multicolumn{3}{c}{\tabincell{c}{Bidirectional\\ RNNs\\128}}&\multicolumn{1}{c}{\tabincell{c}{General\\ RNNs\\128}}\\
    \cmidrule(lr){3-4} \cmidrule(lr){5-7} \cmidrule(lr){8-10} \cmidrule(lr){11-11}&&Paras&Acc.&Paras&\tabincell{c}{speed\\Sec/sample}&Acc.&Paras&\tabincell{c}{speed\\Sec/sample}&Acc.&Acc.\\
    \midrule
    \multirow{4}{*}{\#1.}&
    \rule{0pt}{0.4cm}\tabincell{c}{2 h\_layers}  &1.58mil&92.2\%&0.79mil&0.00127&92.7\%&0.79mil&0.00143&92.7\%&90.6\%\\
    \multirow{4}{*}{}&
    \rule{0pt}{0.4cm}\tabincell{c}{3 h\_layers}  &1.97mil&92.4\%&0.98mil&0.00170&92.9\%&0.98mil&0.00202&92.8\%&91.4\%\\
    \multirow{4}{*}{}&
    \rule{0pt}{0.4cm}\tabincell{c}{4 h\_layers}  &2.37mil&92.6\%&1.18mil&0.00220&92.9\%&1.18mil&0.00270&92.9\%&91.5\%\\
    \multirow{4}{*}{}&
    \rule{0pt}{0.4cm}\tabincell{c}{5 h\_layers}  &2.76mil&92.6\%&1.38mil&0.00284&92.9\%&1.38mil&0.00330&92.9\%&91.5\%\\
    \midrule
    \multirow{2}{*}{\#2.}&
    \rule{0pt}{0.4cm}\tabincell{c}{2 h\_layers}  &1.55mil&95.6\%&0.77mil&0.00113&96.3\%&0.77mil&0.00136&96.1\%&N/A\\
    \multirow{2}{*}{}&
    \rule{0pt}{0.4cm}\tabincell{c}{5 h\_layers}  &2.74mil&95.7\%&1.37mil&0.00231&96.5\%&1.37mil&0.00314&96.4\%&N/A\\
    \bottomrule
    \end{tabular}
    \end{threeparttable}
    \tabincell{cl}{\it 256,128 denotes the hidden layer size of corresponding RNNs. \\h\_layers denotes hidden layers.\\ \#1. and \#2. denote experiments carried out on the IAHCC-UCAS2016 dataset and CASIA-OLHWDB1.1 dataset respectively.}
\end{table*}
\subsection{Data Preprocessing}
the location sequence of a handwritten character is denoted by: $\bm{x}^i=(\bm{x}_1^{(i)},\bm{x}_2^{(i)},\cdots,\bm{x}_t^{(i)},\cdots,\bm{x}_T^{(i)})$,
where the $\bm{x}_t^{(i)}$ denotes $t$th location dot vector, the $\bm{x}_t^{(i)}$ could be represented as $ \bm{x}_t^i=(m_t^i, n_t^i),\ t=1,2,\cdots,T$,
where $m_t^{(i)}$ and $n_t^{(i)}$  denote $t$th location dot's coordinate values respectively. Before being fed into our system, the data is preprocessed by two steps. Concretely,
 (1) $\bm{m}^{(i)}$ $(m_1^{(i)}, m_2^{(i)},\cdots,m_T^{(i)})$ and $\bm{n}^i$ $(n_1^{(i)}, n_2^{(i)},\cdots,n_T^{(i)})$ of all the locations are scaled to a range of 0 to 64;
 (2) the dot locations are further normalized so that their mean equals to zero, i.e., the new coordinate $(\bm{m'}^{(i)},\bm{n'}^{(i)})$ is obtained by $\bm{m'}^{(i)}=\bm{m}^{(i)}-\bar{\bm{m}^{(i)}}$, $\bm{n'}^{(i)}=\bm{n}^{(i)}-\bar{\bm{n}^{(i)}}$, where $\bar{\bm{m}^{(i)}}$ and $\bar{\bm{n}^{(i)}}$ denote the means of the corresponding coordinates of all the dot locations.
\subsection{Network Configuration}
In our experiments, the inputs of the RNN networks are 2-dimensional or 3-dimensional temporal sequences, corresponding to coordinates of dot locations on writing trajectory of in-air handwritten character and handwritten character with pen up-and-down mask on touching scren. The number of neurons in the hidden layers are all 256 except some systems in Table~\ref{table1}. The dropout values are all set to 0.6 and mini-batch size is 256. All the computation is performed by GPUs on TESLA K10.
\subsection{Performance Evaluation of Hybrid-parameters RNNs}
The parameters of the proposed hybrid-parameter RNNs are changed during the iteration at temperal dimension. The proposed hybrid-parameter RNNs system could obtain a higher recognition accuracy with fewer parameters than the general RNN. Compared with bidirectional RNNs, the proposed hybrid-parameter RNN has a faster calculation speed. The experiments are carried out on IAHCC-UCAS2016 dataset.
\par
As shown in Table~\ref{table1}, the proposed hybrid-parameter RNNs with five 128-size hidden layers obtain a higher recognition accuracy than the genral RNN with five 256-size hidden layers. However, the parameter quantity of proposed hybrid-parameter RNNs are as half as the general RNNs. Compared with bidirectional RNNs with calculation speed of 0.00127 second/sample, 0.00170 second/sample, 0.00220 second/sample, and 0.00220 second/sample, the proposed hybrid-parameter RNN obtains competitive recognition accuracy with faster calculation speed of 0.00143 second/sample, 0.00202 second/sample, 0.00270 second/sample and 0.00330 second/sample corresponding four kinds of hidden layer size. The calculation time is shortened by about \textbf{15\%}.
\par
For further evaluations, experiments are also carried out on CASIA-OLHWDB1.1 dataset. In the two different architectures of RNNs containing 2 or 5 hidden layers, compared with the general RNNs, the hybrid-parameter RNNs bring parameter decrease of \textbf{50\%} from 1.55mil to 0.77mil , and \textbf{50\%} from 2.74mil to 1.37mil respectively and obtain high recognition results. Compared with the bidirectional RNNs, the the hybrid-parameter RNNs bring calculation speed increase of \textbf{18\%} from 0.00136 second/sample to 0.00113 second/sample, and \textbf{26\%} from 0.00314 second/sample to 0.00231 second/sample respectively and obtain competitive results.
\subsection{Performance Evaluation of Memory Pool Unit}
\begin{table}[htbp]
\setlength{\abovecaptionskip}{10pt}
\setlength{\belowcaptionskip}{10pt}
\caption{Recognition accuracy of different methods}
  \label{table2}
  \centering
  \begin{tabular}{ccccc}
    \toprule
    \rule{0pt}{0.4cm}Methods &\tabincell{c}{GRU} &\tabincell{c}{LSTM} &\tabincell{c}{MPU}&\tabincell{c}{MPU\&C}\\
    \midrule
    \rule{0pt}{0.4cm}\tabincell{c}{2 h\_layers } &\tabincell{c}{92.2\%}&\tabincell{c}{91.6\%}&\tabincell{c}{92.5\%}&\tabincell{c}{92.5\%}\\
    \rule{0pt}{0.4cm}\tabincell{c}{3 h\_layers} &\tabincell{c}{92.4\%}&\tabincell{c}{92.0\%}&\tabincell{c}{92.7\%}&\tabincell{c}{92.7\%}\\
    \rule{0pt}{0.4cm}\tabincell{c}{4 h\_layers } &\tabincell{c}{92.6\%}&\tabincell{c}{92.2\%}&\tabincell{c}{92.7\%}&\tabincell{c}{92.7\%}\\
    \rule{0pt}{0.4cm}\tabincell{c}{5 h\_layers} &\tabincell{c}{92.6\%}&\tabincell{c}{92.2\%}&\tabincell{c}{92.8\%}&\tabincell{c}{92.8\%}\\
    \bottomrule
  \end{tabular}
\par
\end{table}
The proposed MPU is applied to in-air handwritten Chinese character recognition. The main part of the proposed method is a memory pool which could learn the useful temporal information. And the core of the memory pool is the renewal function. Input compensation is important for network training and two methods are proposed for input compensation. One is that the network inputs are contained in the inputs of each layer. The other method is using the memory pool outputs and last layer outputs to compute the current hidden layer state.
\par
To evaluate the performance of the MPU, experiments are carried out on IAHCC-UCAS2016 dataset. The RNN structures used in these experiments are the general RNNs equipped with GRU ,LSTM, MPU and MPU with input compensation respectively. As shown in Table~\ref{table2}, the MPU and MPU with input compensation methods obtain a competitive recognition accuracy. Compared with LSTM, the proposed method has fewer parameters and a simpler structure. Compared with GRU, the whole hidden layer state calculation process are more reasonable and convictive.
\par
For further performance evaluation of the MPU, another experiment was carried out on CASIA-OLHWDB1.1 dataset~\cite{cit26}. The data from 36 persons in training set are randomly sampled to form the validation set.
As shown in Table~\ref{table3}, in the two different architectures of RNNs containing 2 or 5 hidden layers, compared with the exsiting structure LSTM, the RNNs  equipped with MPU obtains a higher recognition results.
\begin{table}[htbp]
\setlength{\abovecaptionskip}{10pt}
\setlength{\belowcaptionskip}{10pt}
\caption{Recognition accuracy of different methods}
  \label{table3}
  \centering
  \begin{tabular}{cccc}
    \toprule
    \rule{0pt}{0.3cm}Methods &\tabincell{c}{MPU\&C}&\tabincell{c}{GRU}&\tabincell{c}{LSTM} \\
    \midrule
    \rule{0pt}{0.3cm}\tabincell{c}{2 h\_layers } &\tabincell{c}{95.9\%}&\tabincell{c}{95.6\%}&\tabincell{c}{95.3\%}\\
    \rule{0pt}{0.3cm}\tabincell{c}{5 h\_layers } &\tabincell{c}{96.1\%}&\tabincell{c}{95.7\%}&\tabincell{c}{95.4\%}\\
    \bottomrule
  \end{tabular}
\par
\end{table}
\subsection{Performance Evaluation of Stacked Different Hidden Layer States}
\begin{table}[htbp]
\setlength{\abovecaptionskip}{10pt}
\setlength{\belowcaptionskip}{10pt}
\caption{Recognition accuracy of different methods}
  \label{table4}
  \centering
  \begin{tabular}{cccc}
    \toprule
    \rule{0pt}{0.3cm}Methods &\tabincell{c}{with\\ stacked} &\tabincell{c}{without\\ stacked} &\tabincell{c}{synthesize with \\weighted matrix}\\
    \midrule
    \rule{0pt}{0.3cm}\tabincell{c}{2 h\_layers } &\tabincell{c}{92.6\%}&\tabincell{c}{92.2\%}&\tabincell{c}{90.2\%}\\
    \rule{0pt}{0.3cm}\tabincell{c}{3 h\_layers} &\tabincell{c}{92.8\%}&\tabincell{c}{92.4\%}&\tabincell{c}{91.2\%}\\
    \rule{0pt}{0.3cm}\tabincell{c}{4 h\_layers } &\tabincell{c}{92.8\*\%}&\tabincell{c}{92.6\%}&\tabincell{c}{91.8\%}\\
    \rule{0pt}{0.3cm}\tabincell{c}{5 h\_layers} &\tabincell{c}{92.8\%}&\tabincell{c}{92.6\%}&\tabincell{c}{92.0\%}\\
    \bottomrule
  \end{tabular}
\par
\end{table}
For RNNs with same hidden layer size in each layer, we make a suggestion that all the outputs of each layer are stacked as the output of network. By synthesizing all the outputs of each layer, Stacked method increases the recognition accuracy without increasing the parameters.
\par
To evaluate the performance of our proposed stack method, experiments are carried out on IAHCC-UCAS2016 dataset. The RNNs structure used in the four experiments are the general RNNs equipped with GRU. The experimental results are shown in Table~\ref{table4}. The experimental results show that the proposed stack method could bring a recognition accuracy increase.
\subsection{Recognition Accuracy Comparison between Ours and the State-of-arts Method}
Based on the constructed RNNs in Table~\ref{table1} and Table~\ref{table2}, we construct an ensemble classifier containing all the networks. Concretely, the output of the ensemble classifier ${\bm{y}}$ is the sum of outputs of child RNNs, i.e., ${\bm{y}}=\sum_{j=1}^{N}{\bm{y}_j}$, where $\bm{y}_j$ denotes the output of a network in Table~\ref{table1} and Table~\ref{table2}, $N$ denotes N networks in Table~\ref{table1} and Table~\ref{table2}.
\par
Since the accuracy in~\cite{cit24} is obtained only for the 3811 Chinese character classes, the experimental results in Table~\ref{table4} are also reported on the 3811 Chinese characters for the sake of fair comparison.
From Table~\ref{table5}, we can see the ensemble classifier can achieve the recognition accuracy of 93.7\%. The proposed classifiers outperform the state-of-the-art results~\cite{cit24,cit31}.
\begin{table}[htbp]
\setlength{\abovecaptionskip}{10pt}
\setlength{\belowcaptionskip}{10pt}
\caption{Comparison of recognition accuracy between ours and the state-of-the-art method~\cite{cit24}}
  \label{table5}
  \centering
  \begin{tabular}{cccc}
    \toprule
    \rule{0pt}{0.3cm} \tabincell{c}{Method}    &\tabincell{c}{Ours Ensemble}    &\tabincell{c}{Method\#1.}          &\tabincell{c}{Method\#2.}\\
    \midrule
    \rule{0pt}{0.3cm}\tabincell{c}{Acc.}          &93.7\%                                 &93.4\%                                          &91.8\%\\
    \bottomrule
  \end{tabular}
  \tabincell{cl}{\it Method\#1.~\cite{cit31},Method\#2.~\cite{cit24} }
\end{table}
\section{Conclusion}
This paper presents an end-to-end recognizer for online in-air handwritten Chinese characters by using recurrent neural networks (RNN) and it has obtained competitive performance compared with the state-of-the-art methods~\cite{cit31,cit24}.  The merit of the proposed method is that it does not need the explicit feature representation in modeling the classifier. To make the classic RNN work better for online IAHCCR, Three mechanisms are proposed, i.e., the Menory Pool method, hybrid-parameters RNNs, stacked different hidden Layer States. The experimental results show that all the mechanisms can effectively promote the classification performance of RNN.
\section{Acknowledgments}
This paper is under consideration at Pattern
Recognition  Letters.
\newpage
\bibliographystyle{aaai}
\bibliography{refs}

\end{document}